\documentclass[letterpaper]{article} 
\usepackage{aaai2026}  
\usepackage{times}  
\usepackage{helvet}  
\usepackage{courier}  
\usepackage[hyphens]{url}  
\usepackage{graphicx} 
\usepackage{natbib}  
\usepackage{caption} 
\usepackage{algorithm}
\usepackage{algorithmic}

\usepackage{newfloat}
\usepackage{listings}
\usepackage{amssymb}
\usepackage{amsmath}
\usepackage{booktabs}
\usepackage{multirow}
\usepackage{algorithm}
\usepackage{tabularx}
\DeclareCaptionStyle{ruled}{labelfont=normalfont,labelsep=colon,strut=off} 
\floatstyle{ruled}
\newfloat{listing}{tb}{lst}{}
\floatname{listing}{Listing}
\title{CLUE: Leveraging Low-Rank Adaptation to Capture Latent Uncovered Evidence for Image Forgery Localization}
\author {
    Youqi Wang\textsuperscript{\rm 1},
    Shunquan Tan\textsuperscript{\rm 2},
    Rongxuan Peng\textsuperscript{\rm 1},
    Bin Li\textsuperscript{\rm 2},
    Jiwu Huang\textsuperscript{\rm 2}
}
\affiliations {
    \textsuperscript{\rm 1}Guangdong Provincial Key Laboratory of Intelligent Information Processing, Shenzhen University, China\\
    \textsuperscript{\rm 2}Faculty of Engineering, Shenzhen MSU-BIT University, China\\
    2400042020@mails.szu.edu.cn,
    tanshunquan@gmail.com,
    2350433004@email.szu.edu.cn,
    libin@szu.edu.cn,
    jwhuang@smbu.edu.cn
}

\usepackage{bibentry}

\begin{document}

\maketitle

\begin{abstract}
    The increasing accessibility of image editing tools and generative AI has led to a proliferation of visually convincing forgeries, compromising the authenticity of digital media. In this paper, in addition to leveraging distortions from conventional forgeries, we repurpose the mechanism of a state-of-the-art (SOTA) text-to-image synthesis model by exploiting its internal generative process, turning it into a high-fidelity forgery localization tool. To this end, we propose CLUE (Capture Latent Uncovered Evidence), a framework that employs Low-Rank Adaptation (LoRA) to parameter-efficiently reconfigure Stable Diffusion 3 (SD3) as a forensic feature extractor. Our approach begins with the strategic use of SD3's Rectified Flow (RF) mechanism to inject noise at varying intensities into the latent representation, thereby steering the LoRA-tuned denoising process to amplify subtle statistical inconsistencies indicative of a forgery. To complement the latent analysis with high-level semantic context and precise spatial details, our method incorporates contextual features from the image encoder of the Segment Anything Model (SAM), which is parameter-efficiently adapted to better trace the boundaries of forged regions. Extensive evaluations demonstrate CLUE's SOTA generalization performance, significantly outperforming prior methods. Furthermore, CLUE shows superior robustness against common post-processing attacks and Online Social Networks (OSNs). Code is publicly available at https://github.com/SZAISEC/CLUE.
\end{abstract}

\section{Introduction}
\label{sec:introduction}

\par
The advanced capabilities of digital forgery tools empower users to create visually convincing forgeries with unprecedented ease, spanning a wide range of techniques from traditional forgeries like copy-move, splicing, and inpainting \cite{Bappy2017Exploiting} to modern, AI-driven techniques such as high-fidelity generative inpainting \cite{Lugmayr2022RePaintCVPR} and complex language-guided edits \cite{Fu2024GuidingICLR, Brooks2023InstructPix2PixCVPR}. The spread of these deceptive images can fuel disinformation, undermine the reliability of legal evidence, and erode public trust. Consequently, Image Forgery Localization, which aims to precisely localize forged regions, has become an essential field of study for maintaining the trustworthiness of digital images.

\par
While the Image Forgery Localization field has advanced significantly, the practical application of contemporary methods is often hindered by two shortcomings: (i) poor generalization to novel forgery techniques, and (ii) inadequate robustness against common image degradations.

Poor generalization stems from the inherent difficulty of identifying unseen forgeries during training. To mitigate this, one line of research has focused on identifying well-defined, low-level artifacts, such as JPEG compression traces \cite{Kwon2021CATNetWACV, Sheng2024DiRLocTDSC} or noise patterns \cite{Guillaro2023TruForCVPR}. To address specific artifacts, other works have explored multi-view frameworks that create richer feature representations by combining information from disparate domains \cite{Chen2021MVSSNetICCV, Dong2023MVSSNetTPAMI}. Another line of work aims to improve generalization by mitigating semantic biases, for instance, by decoupling content from forgery traces \cite{Sun2023SAFLNetICCV}. However, these approaches are discriminative by design, learning to recognize the statistical artifacts of known forgeries. This reliance on known artifacts is a critical vulnerability, leading to significant performance degradation when confronted with forgeries from novel generative models, which introduce distinct and unforeseen statistical footprints \cite{Kong2025PixelInTPAMI}.

\begin{figure}[t]
    \centering
    \includegraphics[width=1.04\columnwidth]{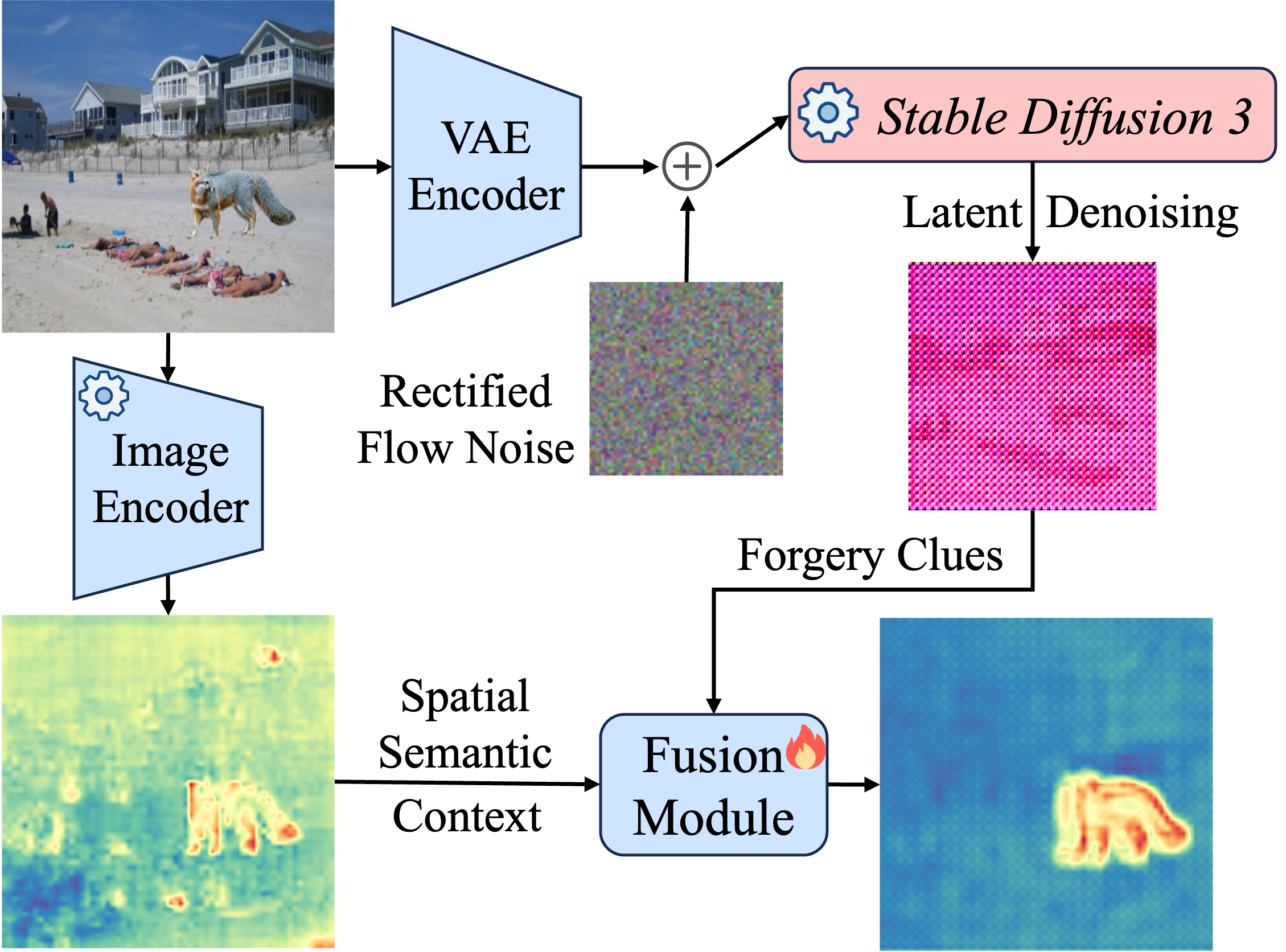} 
    \caption{How RF noise guides a LoRA-tuned SD3 to isolate and amplify hidden forgery traces.}
    \label{fig:mask_comparison}
    \end{figure}

\par
The issue of inadequate robustness presents another critical hurdle. For instance, foundational work in this area employed Photoshop scripting to synthesize more authentic training examples \cite{Zhuang2021Image}, while other research has introduced a Self-Adversarial Training (SAT) regimen to build more resilient models \cite{Zhuo2022Self}. Despite these advances and other effective strategies \cite{Han2024HDFNetTPAMI, Liu2024AttentiveTIFS, Bai2025PIMNetPR, Li2025AdaIFLECCV}, the prevailing paradigm remains artifact-driven. This approach is challenged by the rise of large pre-trained models \cite{Liu2024PromptIMLMM}, whose adaptation for forgery localization often proves complex or suboptimal \cite{Peng2024CoDETIFS, Lou2025MPCTIFS, Sheng2025StereoSRPR}. These limitations motivate a framework that understands the generative principles of a forgery, rather than merely identifying its residual traces.

\par
To address the aforementioned limitations, we introduce CLUE (Capture Latent Uncovered Evidence), a novel framework that shifts the paradigm from analyzing extrinsic forgery artifacts to exploiting the internal generative process of a powerful diffusion model SD3 for forensic analysis. This principle is embodied in our framework, which repurposes a version of SD3 that has been parameter-efficiently adapted via LoRA \cite{Hu2022LoRAICLR}. As visualized in Figure~\ref{fig:mask_comparison}, the analysis process begins by encoding the input image into a latent representation, to which the RF mechanism then introduces controlled noise. Our approach reconfigures the objective of the denoising process, tailoring its training to guide the model to amplify subtle deviations from the learned distribution of natural images for isolating forgery clues. In parallel, a second branch provides crucial spatial-semantic context for this forensic analysis, utilizing the image encoder from the Segment Anything Model (SAM) \cite{Kirillov2023SAMICCV}. Through LoRA, we selectively fine-tune only the query, key, and value (QKV) projection matrices within its attention blocks, heightening its sensitivity to the subtle boundary artifacts and contextual incongruities that often characterize forged regions. The resulting forensic cues and contextual features are then strategically fused to produce the pixel-accurate localization mask.

\par
Our main contributions are summarized as follows:
\begin{itemize}
    \item We propose CLUE, the first framework to repurpose a generative model SD3 specifically for the Image Forgery Localization task, establishing a new and parameter-efficient paradigm for adapting image synthesis models to forensic analysis.
    \item We demonstrate that the RF noise mechanism, when guided by LoRA fine-tuning, acts as a powerful tool to reveal forgery artifacts. By analyzing latent representations across multiple noise levels, CLUE effectively amplifies subtle inconsistencies inherent to forged regions, making them more salient for localization.
    \item Extensive experiments confirm that CLUE achieves new SOTA localization performance across multiple public benchmarks, outperforming prior methods by demonstrating superior generalization against diverse forgery techniques, not only traditional methods like splicing, copy-move, and removal, but also the sophisticated AI-generated forgeries, as well as exceptional robustness to common post-processing scenarios and OSNs.
\end{itemize}

\par
The remainder of this paper proceeds as follows. Section 2 reviews prior work in Image Forgery Localization and the application of foundation models, namely SAM and diffusion models. Section 3 details the proposed methodology. Section 4 presents the experimental setup, datasets, and comparative results. Section 5 concludes the paper and discusses future directions.

\section{Related Work}
\label{sec:related_work}

\subsection{Image Forgery Localization}
Image Forgery Localization aims to produce a pixel-level mask identifying forged regions within an image. A foundational research pillar in this field is the detection of specific forensic artifacts introduced during the forgery process. This line of work targets diverse cues, from camera-specific traces like noise patterns \cite{Guillaro2023TruForCVPR} and JPEG compression artifacts \cite{Kwon2021CATNetWACV, Sheng2024DiRLocTDSC}, to inconsistencies at object boundaries, addressed via multi-view strategies \cite{Chen2021MVSSNetICCV, Dong2023MVSSNetTPAMI}, spatio-channel correlation \cite{Liu2022PSCCNetTCSVT}, or dedicated attentive mechanisms \cite{Li2025AdaIFLECCV, Liu2024AttentiveTIFS}.

Beyond direct artifact detection, Image Forgery Localization research is also characterized by a diversity in learning paradigms and architectural designs. Significant efforts have focused on explicitly modeling broader pixel and content inconsistencies \cite{Han2024HDFNetTPAMI, Bai2025PIMNetPR, Kong2025PixelInTPAMI}, while other work aims to mitigate semantic bias to improve generalization \cite{Sun2023SAFLNetICCV}. This diversity is also evident in the adoption of alternative frameworks like contrastive learning \cite{Lou2025MPCTIFS} and reinforcement learning \cite{Peng2024CoDETIFS}, and in the development of highly specialized solutions for niche scenarios such as forgery in stereo super-resolution images \cite{Sheng2025StereoSRPR}.

More recently, leveraging large pre-trained models has become a prominent trend in Image Forgery Localization, shifting the focus towards adapting foundational knowledge. This line of work explores various ways to harness these models. For instance, some approaches develop parameter-efficient tuning strategies, such as using prompts to guide large models toward identifying forged content \cite{Liu2024PromptIMLMM}. Others directly incorporate features from foundation models like SAM into their pipeline. For example, Su et al. \cite{Su2024NovelIHMMSEC} designed a new feature extractor that fuses features from SAM's encoder with traditional noise features (SRM) to perform localization.

Our work, CLUE, aligns with this recent trend of leveraging large models but is fundamentally distinct in its methodology and core concept. While prior methods typically employ large models as fixed feature extractors, we introduce a synergistic co-adaptation framework. First, we employ a parameter-efficient LoRA strategy to fine-tune two complementary foundation models in concert: SAM and SD3. Second, and most distinctively, we repurpose the internal generative RF mechanism of SD3 as a specialized forensic feature extractor to expose forged regions. This deep, coordinated adaptation of two powerful foundation models marks a significant departure from previous approaches.

\subsection{SAM for Downstream Tasks}

SAM \cite{Kirillov2023SAMICCV} is a foundational model for prompt-guided image segmentation with remarkable zero-shot capabilities. Its powerful Vision Transformer encoder captures rich spatial-semantic information, making it a versatile tool for various downstream tasks.

The advent of SAM has spurred diverse adaptations. For instance, it has been extended to 3D scenes with NeRFs \cite{Cen2023SA3D}, enhanced with logic reasoners for complex text prompts \cite{Shindo2024DeiSAM}, and leveraged as a front-end for feature matching \cite{Zhang2024MESA}. Another key direction is developing fine-tuning strategies to adapt SAM for challenging domains like remote sensing \cite{Liu2025PointSAM} or to prevent catastrophic forgetting \cite{Liu2024SUM}.

In the specific context of Image Forgery Localization, SAM's strong segmentation prior has also been explored. Some works recast the task as a source partitioning problem guided by point prompts \cite{Kwon2025SAFIREAAAI}, while others use SAM's encoder as a fixed feature extractor, fusing its output with noise features \cite{Su2024NovelIHMMSEC}. However, these methods treat SAM as a standalone component, overlooking the potential of fusing its segmentation capabilities with forensic cues from a generative model.

Our approach differs from these works by introducing a framework centered on the coordinated fine-tuning of two complementary foundation models. While prior methods adapt SAM in isolation or use its encoder as a fixed feature extractor, we use LoRA to heighten SAM's sensitivity to subtle forensic traces, such as boundary artifacts and contextual incongruities. Crucially, this adaptation is performed in coordination with a repurposed SD3 model, enabling a novel fusion of SAM's rich spatial-semantic context with the forensic cues amplified by the generative model. This strategy of jointly adapting complementary models is the key novelty that distinguishes our work from existing methods.

\subsection{Diffusion Models in Forensics}
Diffusion models, originating from foundational works like DDPM \cite{Ho2020DDPMNeurIPS} and popularized by architectures like Latent Diffusion Models underlying Stable Diffusion \cite{Rombach2022LatentDiffCVPR}, have revolutionized image generation. Their ability to synthesize highly realistic images and perform complex edits, such as inpainting \cite{Lugmayr2022RePaintCVPR}, also makes them powerful tools for creating sophisticated forgeries. Consequently, understanding and detecting manipulations involving diffusion models is crucial for forensics. Recent research has started exploring diffusion models within Image Forgery Localization. DiffForensics \cite{Yu2024DiffForensicsCVPR}, for example, leverages a diffusion prior learned via self-supervised pre-training to aid localization, focusing on specific image properties. However, directly harnessing the internal mechanisms of cutting-edge diffusion models, such as the RF process in SD3, specifically for revealing forgery artifacts remains less explored. Our approach, CLUE, uniquely pioneers the use of LoRA-tuned SD3, repurposing its RF-based  as a forensic feature extractor to amplify generative inconsistencies, distinct from prior work that primarily uses diffusion models as general priors or for data augmentation.

\section{Methodology}
\label{sec:methodology}

\begin{figure*}[t!]
    \centering
    \includegraphics[width=\textwidth]{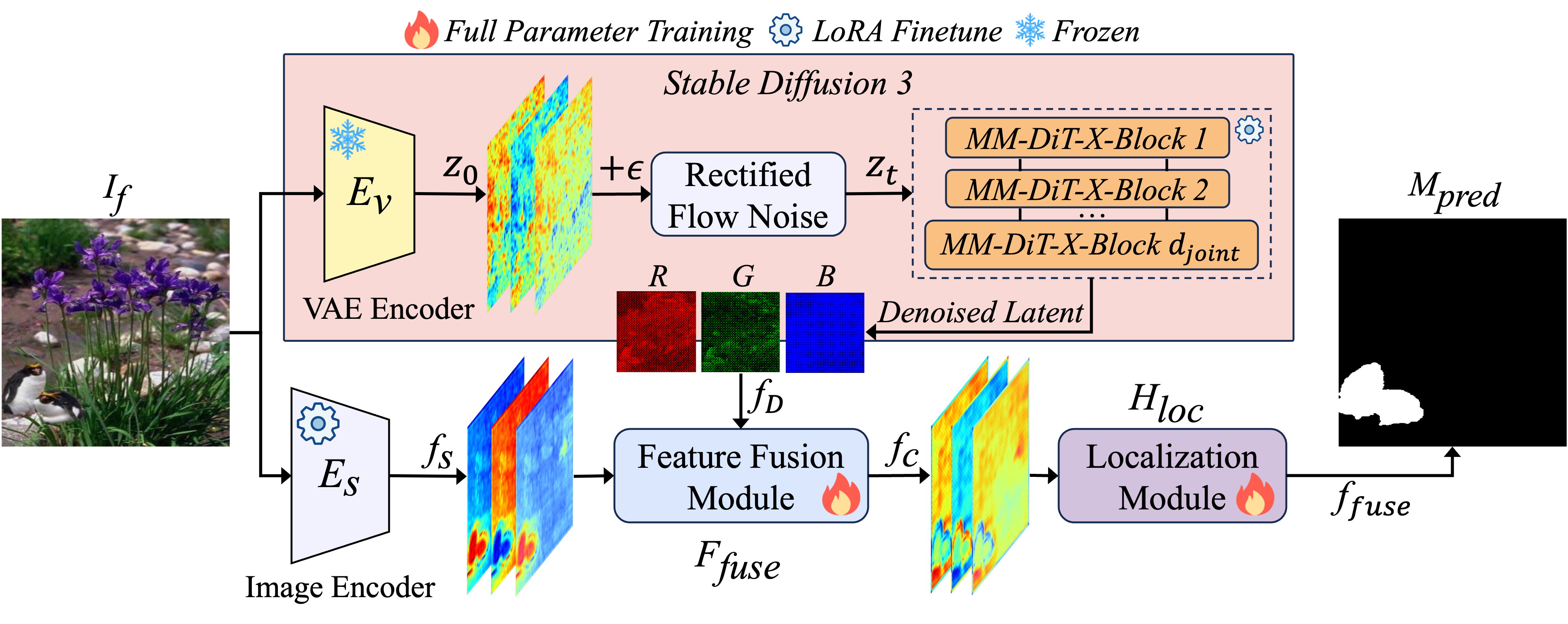}
    \caption{Overall architecture of the proposed CLUE. A forged image $I_f$ is processed by two parallel branches. The top branch uses a frozen VAE Encoder ($E_V$) to obtain latent $z_0$, applies RF noise addition, and feeds the noisy latents $z_t$ into a LoRA-tuned SD3 model acting as a forensic feature extractor to produce features $f_D$. The bottom branch uses a LoRA-tuned SAM Image Encoder ($E_S$) to extract semantic features $f_S$. Both feature sets are then fused by a trainable Feature Fusion module ($F_{fuse}$) and processed by a trainable Localization Module ($H_{loc}$) to generate the final predicted mask $M_{pred}$.}
    \label{fig:overview}
\end{figure*}

\subsection{Preliminaries and Problem Formulation}
The goal of Image Forgery Localization is to identify manipulated regions within a given image $I \in \mathbb{R}^{H_{orig} \times W_{orig} \times C_{in}}$, where $H_{orig}, W_{orig}$ are the original spatial dimensions and $C_{in}$ is the number of input channels (typically 3 for RGB). Specifically, the task involves predicting a binary mask $M_{pred} \in [0, 1]^{H_{orig} \times W_{orig}}$ that accurately segments the forged pixels (label 1) from the authentic background pixels (label 0), aligning closely with the ground truth mask $M_{gt}$. The primary challenge of this task lies in detecting the subtle and diverse traces left by sophisticated forgeries.

\subsection{The CLUE Framework}
The CLUE framework, illustrated in Figure~\ref{fig:overview}, is designed to overcome the critical Image Forgery Localization challenges of generalization and robustness through a strategic dual-branch architecture. The primary branch addresses the generalization challenge by repurposing SD3's generative process into a forensic analysis that operates within the latent space. This process involves perturbing the image's latent representation with RF noise and training the LoRA-tuned SD3 model to amplify statistical inconsistencies. Such a mechanism identifies forgeries based on their fundamental deviation from natural image distributions, enabling effective localization without relying on specific artifact patterns. In parallel, a complementary branch tackles robustness and precision by using a LoRA-tuned image encoder to extract stable spatial-semantic features. This is effective because the SAM encoder is adept at capturing an image's core spatial and semantic structures, which are far more resilient to post-processing degradations like compression and blurring than fine-grained forensic artifacts, thus providing a stable contextual map for accurate boundary delineation. These complementary feature streams are then fused and processed by a Localization Module to yield the final localization mask.

\subsubsection{Spatial-Semantic Feature Extraction}
\label{sec:sam_branch}

In parallel, the SAM branch utilizes the powerful image encoder ($E_S$) from SAM \cite{Kirillov2023SAMICCV}. This encoder provides a strong prior for forgery detection by capturing rich spatial-semantic features that model the natural image regularities often violated by forgeries. We employ LoRA \cite{Hu2022LoRAICLR} to adapt $E_S$ for the specific demands of forgery localization, a parameter-efficient strategy that preserves the encoder's core knowledge. Fine-tuning is applied only to the query, key, and value (QKV) projection matrices within the encoder's attention blocks. Modifying the attention mechanism in this parameter-efficient manner allows $E_S$ to learn sensitivity towards the subtle boundary artifacts and contextual incongruities characteristic of forgeries, effectively refocusing its representational capacity on forensic cues without costly full-model retraining. The output of this LoRA-adapted SAM branch is the feature map $f_S \in \mathbb{R}^{C_S \times H_S \times W_S}$. These features encapsulate both high-level semantic understanding and precise spatial localization information, serving as a rich spatial-semantic context map against which the forensic cues from the SD3 branch can be evaluated in the subsequent fusion stage.

\subsubsection{Generative Inconsistency Extraction}
\label{sec:sd3_branch} 

This branch forms the core of our forensic analysis, repurposing a generative model to extract subtle generative inconsistencies. The process begins by leveraging the frozen VAE encoder ($E_V$) from SD3 to map the input image $I_f$ into an initial latent representation $z_0$:
\begin{equation}
    z_0 = E_V(I_f)
    \label{eq:vae_latent}
\end{equation}
where $z_0 \in \mathbb{R}^{C_z \times H \times W}$. This latent representation is then perturbed using the RF mechanism, which generates a noisy version $z_t$ at time $t \in [0, 1]$ by linearly interpolating between $z_0$ and a standard normal distribution $\epsilon$ \cite{Liu2024InstaFlowICLR}:
\begin{equation}
    z_t = (1 - t) z_0 + t \epsilon
    \label{eq:rf_noise}
\end{equation}
where $\epsilon \sim \mathcal{N}(0, \mathbf{I})$ has the same dimensions as $z_0$.

The parameter-efficient LoRA adaptation reconfigures the objective of the SD3 model as it processes each noisy latent $z_t$. The model is thereby trained to be highly sensitive to deviations from the learned distribution of natural images, compelling it to amplify these generative inconsistencies and yield a feature map, $f_D \in \mathbb{R}^{C_D \times H \times W}$, that highlights the anomalies for the subsequent fusion stage.

\subsubsection{Feature Fusion Module.} 
\label{sec:feature_fusion}

\begin{figure}[h!] 
    \centering
    \includegraphics[width=1.02\columnwidth]{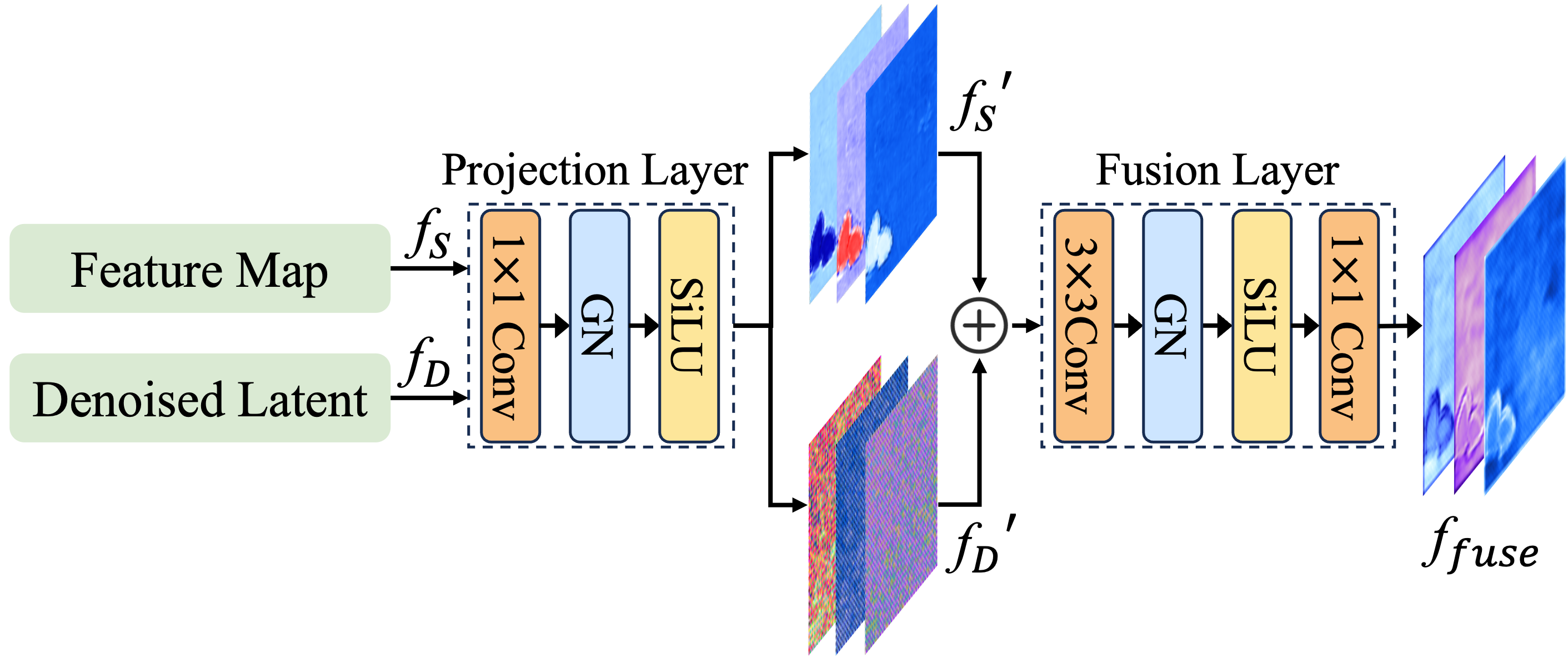} 
    \caption{Architecture of the Feature Fusion module.}
    \label{fig:feature_fusion_diagram} 
\end{figure}

To effectively harness complementary information for robust forgery localization, the semantic characteristics $f_S \in \mathbb{R}^{C_S \times H_S \times W_S}$ of the SAM branch and the consolidated forensic characteristics $f_D \in \mathbb{R}^{C_D \times H \times W}$ of the SD3 branch serve as input to the trainable Feature Fusion module $F_{fuse}$. The specific architecture of this module is illustrated in Figure~\ref{fig:feature_fusion_diagram}.

As shown in the figure, both input feature sets, $f_S$ and $f_D$, undergo parallel processing through independent projection layers to produce intermediate representations, $f'_S$ and $f'_D$. This initial transformation is crucial for aligning the distinct feature spaces of the semantic and forensic cues. These transformed features are then concatenated and passed through a refinement sequence that comprises a $3 \times 3$ convolution, another Group Normalization, a SiLU activation \cite{Elfwing2018SiLU}, and a concluding $1 \times 1$ convolution. This sequence is designed to effectively model the local spatial correlations between the newly combined semantic and forensic features. The entire module $F_{fuse}$ learns to optimally merge these cues, yielding the final fused feature map $f_{fuse} \in \mathbb{R}^{C_{fuse} \times H \times W}$ that is then fed to the Localization Module.

\subsubsection{Localization Module.}
\label{sec:segmentation_head}

Finally, the fused features $f_{fuse} \in \mathbb{R}^{C_{fuse} \times H \times W}$ are processed by the trainable Localization Module $H_{loc}$ to generate the pixel-level forgery mask. This lightweight head first applies a $3 \times 3$ convolution followed by a ReLU activation, mapping the features to an intermediate dimension $C_{mid}$. A subsequent $1 \times 1$ convolution then reduces the channels to 1, producing low-resolution forgeries logits. To restore the spatial dimensions to the original image size, these logits are upsampled using bilinear interpolation with a predetermined fixed scale factor, $S_{up}$, calculated based on the ratio between the input feature map size $(H, W)$ and the target output size. Finally, a Sigmoid activation function $\sigma_{sigmoid}$ yields the probability map:
\begin{equation}
    M_{pred} = \sigma_{sigmoid}({Upsample}(H_{conv}(f_{fuse}) ) )
    \label{eq:seg_head}
\end{equation}
where $H_{conv}$ represents the convolutional layers, \textit{Upsample} denotes the fixed-scale bilinear interpolation, and the output mask is $M_{pred} \in [0, 1]^{1 \times H_{orig} \times W_{orig}}$.

\begin{table*}[tb!]
    \centering
    \small
    \caption{Generalization performance comparison with SOTA methods across benchmarks based on F1/IoU scores (at a fixed threshold). Best and second-best results are in \textbf{bold} and \underline{underlined}.}
    \label{tab:generalization_results}
    
    \setlength{\tabcolsep}{2.845pt}

    \begin{tabular}{@{}ccccccccccccccccccccc@{}}
        \toprule
        \multirow{2}{*}{Method} & \multicolumn{2}{c}{CASIA-v1} & \multicolumn{2}{c}{Columbia} & \multicolumn{2}{c}{NIST16} & \multicolumn{2}{c}{Coverage} & \multicolumn{2}{c}{ACDSee} & \multicolumn{2}{c}{CocoGlide} & \multicolumn{2}{c}{MISD} & \multicolumn{2}{c}{IPM15K} & \multicolumn{2}{c}{AutoSplice} & \multicolumn{2}{c}{Weighted Avg.} \\
        \cmidrule(lr){2-3} \cmidrule(lr){4-5} \cmidrule(lr){6-7} \cmidrule(lr){8-9} \cmidrule(lr){10-11} \cmidrule(lr){12-13} \cmidrule(lr){14-15} \cmidrule(lr){16-17} \cmidrule(lr){18-19} \cmidrule(lr){20-21}
        & F1 & IoU & F1 & IoU & F1 & IoU & F1 & IoU & F1 & IoU & F1 & IoU & F1 & IoU & F1 & IoU & F1 & IoU & F1 & IoU \\
        \midrule
        MVSS-Net & .432 & .379 & .677 & .588 & .305 & .248 & .397 & .335 & .267 & .210 & .333 & .257 & .640 & .511 & .191 & .148 & .291 & .213 & .236 & .184 \\
        IF-OSN & .509 & .465 & .706 & .607 & .268 & .199 & .261 & .191 & .323 & .247 & .264 & .207 & .711 & .586 & .380 & .307 & .508 & .394 & .407 & .328 \\
        TruFor & .696 & .633 & .798 & .740 & .362 & .291 & \underline{.480} & \underline{.411} & \underline{.485} & \underline{.406} & .360 & .292 & .659 & .533 & \underline{.545} & \underline{.465} & \underline{.654} & \textbf{.543} & \underline{.563} & \underline{.478} \\
        CoDE & \underline{.723} & \underline{.637} & .881 & .844 & \underline{.420} & \underline{.339} & .445 & .358 & .465 & .372 & \underline{.489} & .387 & \underline{.760} & \underline{.642} & .540 & .450 & .576 & .443 & .553 & .456 \\
        SAFIRE & .299 & .238 & \underline{.900} & \underline{.884} & .168 & .138 & .253 & .209 & .359 & .284 & .479 & \underline{.404} & .620 & .489 & .285 & .230 & .274 & .196 & .295 & .236 \\
        \textbf{CLUE (Ours)} & \textbf{.847} & \textbf{.795} & \textbf{.938} & \textbf{.917} & \textbf{.591} & \textbf{.519} & \textbf{.673} & \textbf{.610} & \textbf{.617} & \textbf{.536} & \textbf{.706} & \textbf{.613} & \textbf{.766} & \textbf{.647} & \textbf{.745} & \textbf{.690} & \textbf{.683} & \underline{.522} & \textbf{.733} & \textbf{.658} \\
        \bottomrule
    \end{tabular}
\end{table*}
    
\subsection{Training Objective}
The trainable components of CLUE are jointly optimized end-to-end using a weighted sum of Binary Cross-Entropy (BCE) and Dice loss. By directly optimizing for mask overlap, the Dice loss component stabilizes the training process, especially in scenarios with sparse forged regions. In parallel, the BCE loss provides the fine-grained, per-pixel supervision required for the model to precisely localize subtle forgery artifacts.

\begin{equation}
    \label{eq:bce_averaged} 
    L_{BCE} = \frac{1}{N} \sum_{i=1}^{N} l_{bce}(M_{gt, i}, M_{pred, i})
    \end{equation}
    where $l_{bce}(y, \hat{y}) = - [ y \log(\hat{y}) + (1 - y) \log(1 - \hat{y}) ]$ is the per-pixel BCE loss.
    \begin{equation}
    \label{eq:dsc_pure}
    L_{Dice} = \frac{2 \sum_{i=1}^{N} M_{pred, i} M_{gt, i}}{\sum_{i=1}^{N} M_{pred, i} + \sum_{i=1}^{N} M_{gt, i}}
    \end{equation}
where $N$ represents the total number of pixels in the batch, and the summation is over all pixels. $M_{gt, i}$ and $M_{pred, i}$ are the ground truth and predicted values for the $i$-th pixel.
\begin{equation}
\label{eq:total_loss}
    L_{total} = \lambda_{BCE} L_{BCE} + \lambda_{Dice} L_{Dice}
\end{equation}
We typically initialize $\lambda_{BCE}=\lambda_{Dice}=0.5$ and dynamically adjust them according to the validation performance.

\section{Experiments}
\label{sec:experiments}

\subsection{Datasets and Evaluation Metrics}

\paragraph{Training.}
Our training set comprises a total of 34,538 images. The protocol generally follows TruFor \cite{Guillaro2023TruForCVPR}, using a selection of public datasets including CASIAv2 \cite{Dong2013CASIADatabase}, IMD2020 \cite{Novozamsky2020IMD2020}, FantasticReality \cite{Kniaz2019PointFantasticReality}, and TampCOCO \cite{Kwon2022LearningJPEGArtiIJCV}. Differing from the TruFor setup, our selection utilizes only a 1\% subset of TampCOCO and excludes the TampRAISE dataset. Further details on the datasets are provided in the Appendix.

\paragraph{Testing.}
To evaluate generalization, CLUE is tested on a diverse collection of nine benchmarks, comprising a total of 21,535 images. This set includes classic datasets for traditional forgeries (CASIA-v1 \cite{Dong2013CASIADatabase}, Columbia \cite{Ng2004DataSetTR}, NIST16 \cite{Guan2019MFCDatasetsWACVW}, and Coverage \cite{Wen2016COVERAGEDatabaseICIP}); a benchmark for multi-splicing forgeries (MISD \cite{Kadam2021MISDData}); three modern datasets of AI-generated forgeries (CocoGlide \cite{Jia2023AutoSpliceData}, IPM15K \cite{Ren2024MFINetTCSVT}, and AutoSplice \cite{Jia2023AutoSpliceData}); and our newly introduced ACDSee \cite{Sun2025} dataset. The ACDSee test set, comprising 337 images, is particularly challenging as it was manually crafted by researchers to simulate realistic scenarios, focusing on the three traditional forgery types of splicing, copy-move, and removal. This diverse suite is chosen to rigorously test generalization across both traditional and sophisticated AI-generated forgery types.

\paragraph{Evaluation Metrics.}
Following the evaluation protocol of MVSS-Net \cite{Dong2023MVSSNetTPAMI}, we primarily use the F1-score (F1) and Intersection over Union (IoU) at the pixel level to quantify the localization performance. Higher values for both metrics indicate better localization accuracy. For fair comparison, all metrics are computed at a fixed binarization threshold of 0.5.

\subsection{State-of-the-art Comparison}

\subsubsection{Generalization Performance.}
For a comprehensive assessment of generalization performance, CLUE was evaluated against several leading methods on testing datasets. Crucially, the training datasets used for all evaluated models, including CLUE, are disjoint from these common testing datasets to ensure a fair comparison. The detailed pixel-level localization results are presented in Table~\ref{tab:generalization_results}.

As shown in Table~\ref{tab:generalization_results}, CLUE achieves the highest weighted average F1 score, significantly outperforming leading methods such as MVSS-Net~\cite{Dong2023MVSSNetTPAMI}, IF-OSN ~\cite{Wu2022IF-OSN}, TruFor~\cite{Guillaro2023TruForCVPR}, CoDE~\cite{Peng2024CoDETIFS}, and SAFIRE~\cite{Kwon2025SAFIREAAAI} by 0.497$\uparrow$, 0.326$\uparrow$, 0.170$\uparrow$, 0.180$\uparrow$, and 0.438$\uparrow$ respectively. This strong generalization is evident across a comprehensive suite of benchmarks. CLUE secures top performance on benchmarks featuring modern AI-generated forgeries (CocoGlide, IPM15K, and AutoSplice), on the challenging ACDSee test set of realistic, manually crafted forgeries, and on other classic datasets including CASIA-v1, Columbia, NIST16, and MISD. Such leading results across this varied collection of benchmarks confirm its state-of-the-art generalization and localization capabilities.

\begin{figure}[t!]
    \centering
    \includegraphics[width=1\columnwidth]{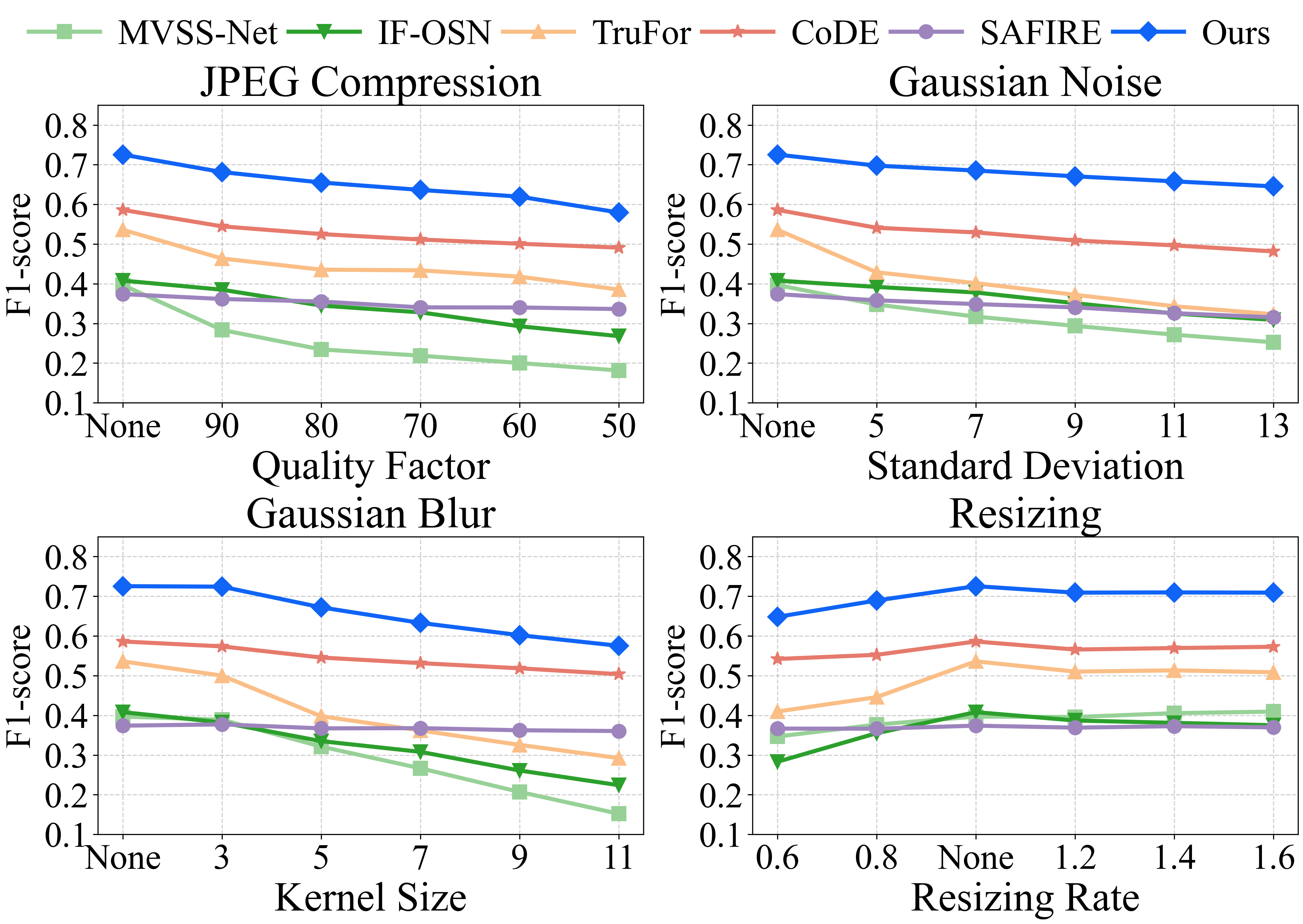} 
    \caption{Robustness comparison based on F1 under common post-processing attacks: JPEG Compression, Gaussian Noise, Gaussian Blur, and Resizing.}
    \label{fig:robustness_analysis}
\end{figure}

\begin{table*}[htbp!]
    \centering
    \small
    \caption{Pixel-level F1 and IoU (at a fixed threshold) performance on datasets processed by various OSNs: Facebook (Fb), WhatsApp (Wa), Weibo (Wb), and WeChat (Wc).}
    \label{tab:osn_robustness}
    
    \setlength{\tabcolsep}{3.5pt}
    
    \begin{tabular}{@{}c cccc cccc cccc cccc cccc@{}}
    \toprule
    \multirow{2}{*}{Method} & \multicolumn{4}{c}{CASIA v1} & \multicolumn{4}{c}{Columbia} & \multicolumn{4}{c}{DSO-1} & \multicolumn{4}{c}{NIST16} & \multicolumn{4}{c}{Weighted Avg.} \\
    \cmidrule(l){2-5} \cmidrule(l){6-9} \cmidrule(l){10-13} \cmidrule(l){14-17} \cmidrule(l){18-21}
     & Fb & Wa & Wb & Wc & Fb & Wa & Wb & Wc & Fb & Wa & Wb & Wc & Fb & Wa & Wb & Wc & Fb & Wa & Wb & Wc \\
    \midrule
    MVSS-Net & .367 & .337 & .389 & .245 & .692 & .687 & .677 & .684 & .263 & .170 & .242 & .205 & .270 & .169 & .248 & .218 & .359 & .305 & .361 & .274 \\
    IF-OSN & .464 & .477 & .465 & .405 & .712 & .726 & .723 & .726 & .403 & .359 & .374 & .367 & .331 & .314 & .295 & .288 & .440 & .440 & .429 & .395 \\
    TruFor & .662 & .669 & .630 & .592 & .764 & .763 & .800 & .773 & .665 & .400 & .478 & .411 & .343 & .400 & .309 & .348 & .569 & .575 & .533 & .520 \\
    CoDE & .699 & .697 & .702 & .629 & .882 & .883 & .884 & .878 & .391 & .377 & .381 & .368 & .412 & .420 & .408 & .413 & .605 & .606 & .605 & .567 \\
    SAFIRE & .266 & .261 & .296 & .215 & .762 & .795 & .862 & .860 & .446 & .303 & .368 & .310 & .426 & .383 & .383 & .360 & .374 & .352 & .380 & .326 \\
    \textbf{CLUE (Ours)} & \textbf{.838} & \textbf{.832} & \textbf{.828} & \textbf{.785} & \textbf{.932} & \textbf{.928} & \textbf{.946} & \textbf{.938} & \textbf{.763} & \textbf{.713} & \textbf{.772} & \textbf{.730} & \textbf{.596} & \textbf{.584} & \textbf{.590} & \textbf{.566} & \textbf{.764} & \textbf{.754} & \textbf{.759} & \textbf{.725} \\
    \bottomrule
    \end{tabular}
    \end{table*}
\subsubsection{Robustness to Post-Processing Attacks.}
Real-world forged images often undergo various post-processing operations that can obscure forgery artifacts. To evaluate CLUE's resilience against these attacks, we subjected images from all datasets utilized in our generalization assessment to post-processing attacks with varying intensities. 

As illustrated in Figure~\ref{fig:robustness_analysis}, CLUE consistently outperforms the competing methods across all post-processing attacks. While the performance of all methods, including CLUE, tends to degrade as the attack intensity increases, CLUE maintains a substantial localization performance lead throughout. This consistent superiority indicates that CLUE's localization capability is significantly more resilient to post-processing attacks compared to others.

Furthermore, we evaluate CLUE's robustness against OSNs. As shown in Table~\ref{tab:osn_robustness}, CLUE consistently outperforms all competing methods. This superiority is most pronounced on WeChat (Wc), which typically applies the most aggressive compression. On this platform, CLUE achieves absolute F1 gains (relative \% in parentheses) of 0.451$\uparrow$ (164.6\%), 0.330$\uparrow$ (83.5\%), 0.205$\uparrow$ (39.4\%), 0.158$\uparrow$ (27.9\%), and 0.399$\uparrow$ (122.4\%) over MVSS-Net, IF-OSN, TruFor, CoDE, and SAFIRE, respectively. These results strongly demonstrate CLUE's superior robustness in real-world scenarios involving OSNs.

\subsection{Ablation Studies}
This section investigates the effects of key components within our framework. Specifically, our analysis focuses on the LoRA fine-tuning for the SAM image encoder, the effectiveness of the SD3 branch and noise mechanism. The presented tables primarily report F1 and IoU on several representative benchmarks.

\begin{table}[htbp!]
    \centering
    \small
    \caption{Ablation on SAM/SD3 components. \protect\checkmark{} denotes LoRA Fine-tuned; \protect\textendash{} denotes Frozen Weights; \protect\texttimes{} denotes Component Removed.}
    \label{tab:ablation_components}
    \setlength{\tabcolsep}{5pt}
    \begin{tabular}{@{}cc cc cc cc@{}}
    \toprule
    \multicolumn{2}{c}{Module Configuration} & \multicolumn{2}{c}{CASIA v1} & \multicolumn{2}{c}{CocoGlide} & \multicolumn{2}{c}{MISD} \\
    \cmidrule(r){1-2} \cmidrule(lr){3-4} \cmidrule(lr){5-6} \cmidrule(l){7-8}
    SD3 & SAM & F1      & IoU     & F1      & IoU     & F1      & IoU     \\ 
    \midrule
    \checkmark & \textendash & .477  & .388  & .415  & .308  & .492  & .350  \\ 
    \checkmark & \texttimes  & .701  & .653  & .432  & .321  & .686  & .552  \\ 
    \texttimes & \checkmark & .745  & .697  & .298  & .232  & .703  & .574  \\ 
    \textendash & \checkmark & .805  & .764  & .564  & .482  & .661  & .531  \\ 
    \textbf{\checkmark} & \textbf{\checkmark} & \textbf{.847} & \textbf{.795} & \textbf{.706} & \textbf{.613} & \textbf{.766} & \textbf{.647} \\
    \bottomrule
    \end{tabular}
\end{table}

\begin{figure*}[t!]
    \centering
    \includegraphics[width=\textwidth, keepaspectratio]{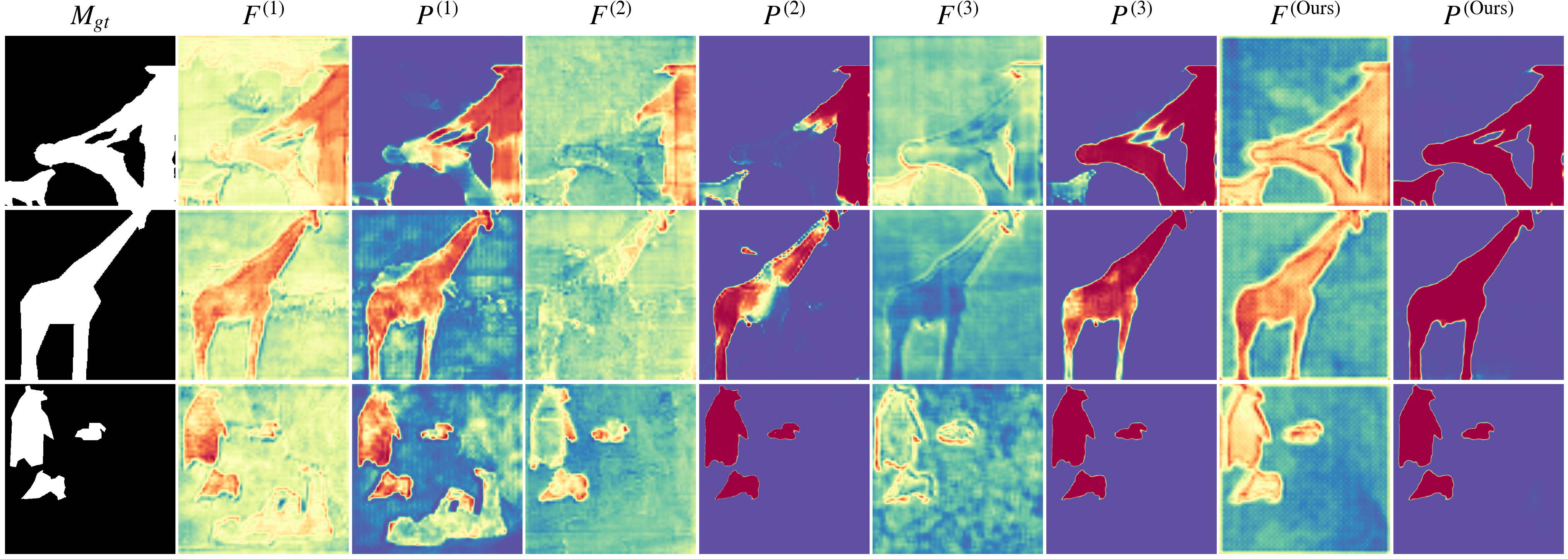}
    \caption{Visual comparison of key ablation study configurations. $M_{gt}$ represents the ground truth mask. The superscripts correspond to four models: (1) with a frozen SAM Image Encoder, (2) with the SD3 branch removed, (3) with a frozen SD3 branch, and (Ours) our full model. $F$: Fused feature map; $P$: Localization probability map.}
    \label{fig:ablation_visual_summary}
\end{figure*}

\subsubsection{Effects of Core Component Configurations.}
Table~\ref{tab:ablation_components} presents an ablation study of our core components. Comparing the full model (5th row) to configurations with a frozen SD3 branch (4th row) or SAM branch (1st row) demonstrates that fine-tuning both is critical. Furthermore, comparing against variants lacking the SD3 branch (3rd row) or the SAM branch (2nd row) confirms the vital contribution of each. These results validate our design of jointly adapting both branches for the Image Forgery Localization task.
    
\begin{table}[htbp!]
    \centering
    \small
    \caption{Ablation study on the noise mechanism in SD3.}
    \label{tab:ablation_noise_mechanism}
    \setlength{\tabcolsep}{5.5pt}
    \begin{tabular}{@{}c cc cc cc@{}}
    \toprule
    \multirow{2}{*}{Noise Mechanism} & \multicolumn{2}{c}{CASIA v1} & \multicolumn{2}{c}{CocoGlide} & \multicolumn{2}{c}{MISD} \\
    \cmidrule(lr){2-3} \cmidrule(lr){4-5} \cmidrule(l){6-7}
                            & F1      & IoU     & F1      & IoU     & F1      & IoU     \\ \midrule
    Zero Noise & .849 & .803 & .647  & .553  & .738  & .612  \\
    DDPM Noise    & .819 & .775 & .524  & .443  & .737  & .614  \\
    \textbf{Rectified Flow Noise} & \textbf{.847} & \textbf{.795} & \textbf{.706} & \textbf{.613} & \textbf{.766} & \textbf{.647} \\
    \bottomrule
    \end{tabular}
\end{table}

Beyond the quantitative metrics in Table~\ref{tab:ablation_components}, Figure~\ref{fig:ablation_visual_summary} offers qualitative insights into our key design choices. With a frozen SAM encoder (1), the prediction $P^{(1)}$ is biased by semantics and fails to precisely isolate the forged region. Removing the SD3 branch entirely (2) eliminates forensic artifact perception, causing the prediction $P^{(2)}$ to completely miss the forgery. Using a frozen SD3 branch (3) allows the model to capture some forensic signals, but the resulting mask $P^{(3)}$ still lacks sharp boundaries. In contrast, CLUE jointly fine-tunes both branches, produces an accurate localization mask $P^{(Ours)}$, confirming that this coordinated adaptation is indispensable for high-quality localization.

\subsubsection{Effects of Noise Mechanism Strategy.}
We evaluate three noise mechanisms within the SD3 branch, with results presented in Table~\ref{tab:ablation_noise_mechanism}. The comparison clearly shows that RF noise mechanism (3rd row) significantly surpasses both a DDPM-based noise strategy (2nd row) and the zero-noise configuration (1st row). This demonstrates that leveraging the structured noise schedule of RF is a more effective strategy for guiding the model to extract discriminative forensic features, making it an integral component of CLUE.

\subsubsection{Effects of RF Parameter: \textit{shift (s)}.}
In this part, we analyze the impact of the RF \textit{shift} parameter ($s$), which controls the noise schedule's intensity. As larger $s$ values raise the minimum noise level, they can affect the model's sensitivity to forgeries with subtle traces.

\begin{table}[htbp!]
    \centering
    \small
    \caption{Ablation study on the RF noise \textit{shift} parameter (\textit{s}).}
    \label{tab:ablation_shift_param}
    \begin{tabular}{@{}c cc cc cc@{}}
        \toprule
        \multirow{2}{*}{\textit{Shift Parameter (s)}} & \multicolumn{2}{c}{CASIA v1} & \multicolumn{2}{c}{CocoGlide} & \multicolumn{2}{c}{MISD} \\
        \cmidrule(lr){2-3} \cmidrule(lr){4-5} \cmidrule(l){6-7}
        & F1      & IoU     & F1      & IoU     & F1      & IoU     \\ \midrule
        0.5 & .815 & .772 & .588 & .495 & .636 & .517 \\
        1.0 & .816 & .774 & .556 & .473 & .680 & .549 \\
        \textbf{3.0} & \textbf{.847} & \textbf{.795} & \textbf{.706} & \textbf{.613} & \textbf{.766} & \textbf{.647} \\
        4.0 & .809 & .766 & .611 & .524 & .611 & .480 \\
        6.0 & .790 & .745 & .615 & .522 & .646 & .527 \\
        \bottomrule
        \end{tabular}
\end{table}

To optimize the \textit{shift (s)} parameter for improved model efficacy, we conducted ablation experiments on several values utilized within SD3. As shown in Table~\ref{tab:ablation_shift_param}, an intermediate shift value yields superior results, with our findings indicating that the shift parameter s = 3.0 consistently achieves the best localization accuracy across the evaluated datasets. Consequently, we select \textit{s} = 3.0 as the \textit{shift} parameter for RF noise mechanism.

\section{Conclusion}
\label{sec:conclusion}
This paper presents CLUE, a novel framework for image forgery localization that employs parameter-efficient LoRA to fine-tune SAM Image Encoder and SD3 in a coordinated manner. CLUE's distinct methodology involves leveraging RF mechanism to analyze latents across varied noise intensities, which are then integrated with features from an adapted SAM. Experimental evaluations demonstrate that CLUE's generalization in forgery localization across various benchmarks, and its robustness against OSNs and common post-processing attacks, surpass those of existing SOTA methods.

Future work includes extending CLUE to a multimodal framework that utilizes SD3's text encoder for query-based forgery localization, optimizing the model's computational efficiency for real-world deployment, and exploring SD3's internal mechanisms to uncover more subtle forensic artifacts. Collectively, these efforts are aimed at pushing the boundaries of image forgery localization field.

\bibliography{aaai2026}

\end{document}